%% file: template.tex
\documentclass{article}
\usepackage[preprint]{colm2026_conference}
\usepackage{microtype}
\usepackage{hyperref}
\usepackage{url}
\usepackage{booktabs}
\usepackage{graphicx}
\usepackage{amsmath,amssymb}
\usepackage{subcaption}
\usepackage{pifont}
\definecolor{darkblue}{rgb}{0, 0, 0.5}
\hypersetup{colorlinks=true, citecolor=darkblue, linkcolor=darkblue, urlcolor=darkblue}

\newcommand{\cmark}{\ding{51}}
\newcommand{\xmark}{\ding{55}}

\title{TraceICL: Thread-Resolved Demonstration Curation with Consistency Calibration for Robust Multimodal In-Context Learning}

\author{Yigit Yilmaz, Elena Petrova, Mehmet Kaya, Lucia Rossi, Amir Rahman \\
Bandirma Onyedi Eylul University \\
\texttt{amir.rahman@bandirma.edu.tr}
}

\begin{document}
\ifcolmsubmission
\fi
\maketitle

\begin{abstract}
Interleaved multimodal in-context learning (ICL) lets frozen large vision--language models (LVLMs) acquire new tasks from a handful of image--text--answer demonstrations, yet it remains fragile: demonstrations bind to the wrong images, redundant exemplars crowd out informative ones, and noisy answers are copied verbatim into predictions. We present \textbf{TraceICL}, a training-free framework that organizes the demonstration pipeline of multimodal ICL into three composable stages. A Thread-Resolved Decomposer (TRD) segments the interleaved context into per-sample threads and re-stitches them with explicit boundaries, restoring image--text binding. An Influence-scored Demonstration Selector (IDS) estimates each candidate's marginal influence with one extra forward pass, removes redundancy, and orders the survivors by descending value. A Weak-to-Strong Consistency Calibration (WSC) module scores each demonstration by a weak sibling model's reconstruction confidence and softly reweights its contribution during decoding. On five vision--language benchmarks with interleaved 4-shot prompts, TraceICL improves the strongest same-setting baseline by $+1.7$ points on average (up to $+2.4$ on VizWiz), shrinks the require-visual performance gap on ScienceQA-IMG, and retains most of its gain under $30\%$ noisy demonstrations, at a $26\%$ latency overhead that an offline cache amortizes. No gradients are ever computed; the LVLM stays frozen.
\end{abstract}

\input{main}

\bibliography{references}
\bibliographystyle{colm2026_conference}

\end{document}

%% file: main.tex
\section{Introduction}

In-context learning (ICL) has become the dominant paradigm for adapting large vision--language models (LVLMs) to new tasks without any parameter update: the prompt interleaves a handful of image--text--answer demonstrations followed by a query, and the model imitates the pattern \citep{zhou2024visual,alayrac2022flamingo}. The paradigm is attractive precisely because demonstrations act as a soft, editable task specification. Yet the same mechanism makes ICL fragile in ways that parameter-efficient fine-tuning is not: every property of the demonstration sequence---its length, order, image--text pairing, and annotation quality---leaks directly into the prediction. Empirical studies of prompting regimes show that seemingly innocuous changes in how context is staged can flip model reliability and even induce hallucinated content \citep{zhao2026deployment}. The problem is amplified in safety-critical multimodal applications: driving world models and video simulators consume interleaved image--text trajectories where a single mis-bound demonstration propagates into an unsafe prediction \citep{li2025driverse}, and perceptual localization stacks likewise couple multiple visual sources that must be attributed correctly \citep{li2025u}.

Existing remedies fall into two families, and each leaves a gap. \emph{Model-side} approaches retrain or restructure the LVLM so that multi-image prompts are handled more gracefully, at the cost of a full instruction-tuning pipeline \citep{zhao2024mmicl,li2023otter}. \emph{Data-side} approaches curate better demonstrations, but selection criteria are usually borrowed from training-time alignment, where influence is measured with gradients and token-level supervision \citep{si2025gateau}, or from plain query--demonstration similarity, which ignores interactions among demonstrations. None of these methods treats the demonstration \emph{sequence} as a first-class object: they do not model how interleaved demonstrations bind to each other, how much marginal information each adds, or how much a noisy demonstration should be trusted.

We address these gaps with \textbf{TraceICL}, a training-free framework that organizes the entire demonstration pipeline of multimodal ICL into three composable stages. First, a \emph{Thread-Resolved Decomposer} (TRD) segments the interleaved context into per-sample threads and re-stitches them with explicit boundaries, turning a chaotic shared stream into locally bound units; the design transfers a thread-decomposition insight originally developed for chaotic long-text reasoning. Second, an \emph{Influence-scored Demonstration Selector} (IDS) estimates the marginal influence of each candidate demonstration with a single extra forward pass, removes redundancy, and orders the survivors by descending value. Third, a \emph{Weak-to-Strong Consistency Calibration} (WSC) module asks a weak sibling model to reconstruct each demonstration's answer and uses the reconstruction confidence to softly reweight the demonstration's contribution during decoding. TraceICL touches only the prompt and the decoding logits; the LVLM stays frozen.

Our contributions are:
\begin{itemize}
\item We identify \emph{cross-demonstration binding}, \emph{demonstration redundancy}, and \emph{unweighted noise injection} as three compounding failure modes of interleaved multimodal ICL, and trace each to a measurable deficiency of current prompting pipelines.
\item We propose TraceICL, the first training-free framework that jointly performs thread-resolved context decomposition, influence-scored demonstration selection, and weak-to-strong consistency calibration for frozen LVLMs.
\item On five vision--language benchmarks with interleaved 4-shot prompts, TraceICL improves the strongest same-setting baseline by $+1.7$ points on average (up to $+2.4$ on VizWiz), reduces the require-visual performance gap on ScienceQA-IMG, and retains most of its gain under $30\%$ noisy demonstrations.
\item We provide a detailed ablation, hyperparameter, robustness, and cost analysis showing that the three modules are complementary and that the overhead is a single extra forward pass plus an offline calibration cache.
\end{itemize}

The rest of this paper reviews related work in Section~\ref{sec:related} on multimodal ICL (\S\ref{sec:rw-mmicl}), demonstration selection (\S\ref{sec:rw-selection}), and supervision reliability (\S\ref{sec:rw-alignment}); formalizes the problem in Section~\ref{sec:method} (\S\ref{sec:formulation}) and details TRD (\S\ref{sec:trd}), IDS (\S\ref{sec:ids}), and WSC (\S\ref{sec:wsc}) with complexity (\S\ref{sec:complexity}); reports experiments in Section~\ref{sec:exp} from setup (\S\ref{sec:setup}) through main results (\S\ref{sec:main}), ablations (\S\ref{sec:ablation}), hyperparameters (\S\ref{sec:hyper}), robustness (\S\ref{sec:robust}), cost (\S\ref{sec:cost}), and limitations (\S\ref{sec:limitation}); and concludes in \S\ref{sec:conclusion}.

\section{Related Work}
\label{sec:related}

\subsection{Multimodal In-Context Learning for LVLMs}
\label{sec:rw-mmicl}

Flamingo established few-shot multimodal ICL with interleaved image--text prompts \citep{alayrac2022flamingo}, and OpenFlamingo made the recipe reproducible \citep{awadalla2023openflamingo}. Follow-up models improve the backbone: BLIP-2 bridges frozen image encoders and LLMs with a lightweight querying transformer \citep{li2023blip2}, InstructBLIP adds instruction tuning \citep{dai2023instructblip}, Otter extends multi-modal instruction following \citep{li2023otter}, and Shikra handles referential dialogue \citep{chen2023shikra}. MMICL shows that the context \emph{scheme}---declarations, templates, and stage-wise formatting---matters as much as the backbone \citep{zhao2024mmicl}. Recent 2026 work continues along this axis: MCD distills multimodal ICL causally \citep{li2026mcd}, inductive--deductive reasoning is injected into demonstrations \citep{wang2026inductive}, parallel ICL reorganizes prompts to shorten effective context \citep{yamaguchi2026parallel}, and ``in-context collapse'' is identified as a characteristic failure of VLMs \citep{rostami2026collapse}. Our work is orthogonal to all of these: we do not change the model, the instruction format, or the reasoning style, but organize \emph{which} demonstrations enter the prompt and \emph{how} they are structured and weighted.

\subsection{Demonstration Selection and Curation}
\label{sec:rw-selection}

In text-only ICL, retrieval-based selection improves over fixed exemplars \citep{li-etal-2023-unified}, active example selection reduces labeling cost while lifting accuracy \citep{zhang-etal-2022-active}, and information-compression views unify selection and ordering \citep{wu-etal-2023-self}. For LVLMs, GATEAU selects influential samples for long-context alignment, but its influence measure requires gradients and token-level annotations \citep{si2025gateau}; counterfactual retrieval \citep{xiong2026counterfactual} and geometry-aware uncertainty coresets \citep{erick2026coresets} improve visual ICL in specific domains, and confusion-aware ICL targets demonstration ambiguity in robotic manipulation \citep{he2026confusion}. These methods optimize either selection \emph{or} context structure; TraceICL couples the two, and additionally weights the selected demonstrations by trustworthiness at decoding time rather than treating them as equally reliable.

\subsection{Reliability and Alignment of Supervision Signals}
\label{sec:rw-alignment}

Weak-to-strong generalization studies how weak supervision can elicit strong capabilities \citep{burns2023weak}, and multi-capability analyses extend the picture to LVLMs \citep{zhou2025weak}. Synthetic tasks with reinforcement learning teach faithfulness to context \citep{DBLP:conf/aaai/SiZGBWGLLHCQZCS26}, yet reward design itself can fail in sparse structured settings and requires diagnostic refinement \citep{wang2026llmreward}, and global planner training shows the value of planning structure over long horizons \citep{si2026goal}. On the reliability side, object hallucination remains pervasive in LVLMs \citep{li-etal-2023-evaluating}, evaluation benchmarks such as MME expose both perception and cognition gaps \citep{fu2023mme}, and even a single poisoned image can jailbreak a VLM, underlining that demonstration-level trust signals are a security concern \citep{tao-etal-2025-imgtrojan}. Fair-comparison studies caution that few-shot fine-tuning and ICL are often conflated \citep{mosbach-etal-2023-shot}. TraceICL imports the weak-to-strong idea into the ICL prompt: the weak model's reconstruction consistency acts as a per-demonstration trust score, avoiding hard filtering that would amplify selection bias.

\section{Method}
\label{sec:method}

\subsection{Problem Formulation}
\label{sec:formulation}

\begin{figure*}[!t]
\centering
\includegraphics[width=\linewidth]{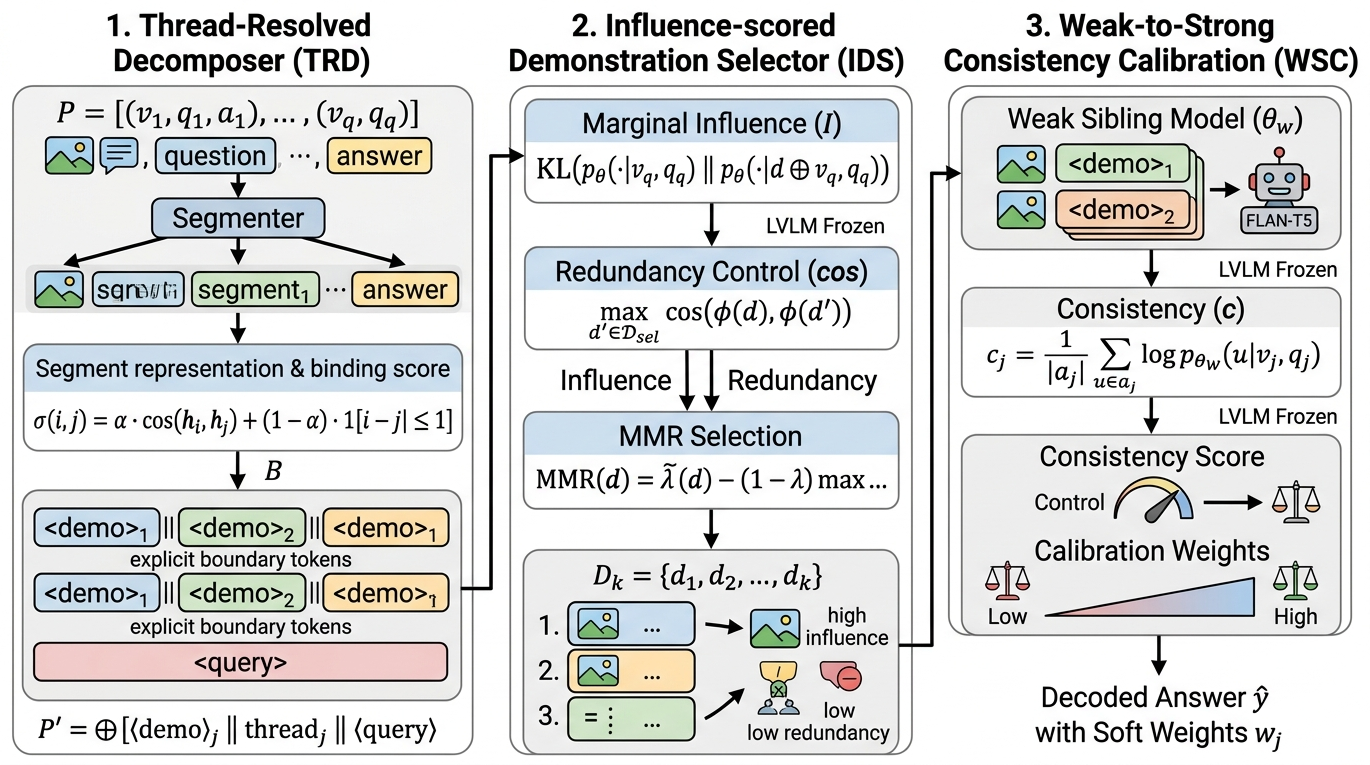}
\caption{Overview of TraceICL. A candidate demonstration pool is filtered by the Influence-scored Demonstration Selector (IDS), organized into per-sample threads by the Thread-Resolved Decomposer (TRD), and softly reweighted at decoding time by the Weak-to-Strong Consistency Calibration (WSC). The LVLM remains frozen throughout.}
\label{fig:arch}
\end{figure*}

Given a frozen LVLM $\theta$, a query sample $(v_q, q_q)$, and a candidate pool $\mathcal{D}$ of demonstrations $(v, q, a)$, multimodal ICL constructs an interleaved prompt $P$ of $k$ demonstrations and decodes the answer:
\begin{align}
P &= \big[(v_1, q_1, a_1),\, \dots,\, (v_k, q_k, a_k),\, (v_q, q_q)\big], \label{eq:prompt} \\
\hat{y} &= \arg\max_{y}\; p_\theta\!\left(y \mid P\right). \label{eq:icl}
\end{align}
The goal of TraceICL is to choose the subset $D_k \subseteq \mathcal{D}$, its thread structure, and a per-demonstration trust weight, so that $p_\theta(y \mid P')$ in \eqref{eq:icl} is maximized \emph{without} updating $\theta$. The full pipeline is shown in Figure~\ref{fig:arch}.

\subsection{Thread-Resolved Decomposer (TRD)}
\label{sec:trd}

Interleaved prompts concatenate image--text segments from different demonstrations into one stream. Attention has no explicit signal for which image belongs to which demonstration, so binding errors grow with sequence length. \citet{zhou2023thread} showed for long text that splitting a chaotic context into ``threads'' and reasoning thread-by-thread restores coherence; we transplant this idea to multimodal prompts.

\textbf{Segmentation and representation.}\quad Let $s_1, \dots, s_m$ be the contiguous image/text segments of the raw prompt of \eqref{eq:prompt}, each encoded as
\begin{align}
h_i = g(s_i), \qquad i = 1, \dots, m,
\end{align}
where $g$ is the frozen encoder (text segments through the LLM embedding, image segments through the vision encoder, with CLIP-style alignment for cross-modal similarity \citep{radford2021learning}). Two segments belong to the same thread if they are semantically adjacent:
\begin{align}
\sigma(i, j) = \alpha \cdot \cos\!\left(h_i, h_j\right) + (1 - \alpha)\cdot \mathbb{1}\!\left[\,\lvert i - j\rvert \le 1\,\right], \label{eq:sigma}
\end{align}
with $\alpha$ in \eqref{eq:sigma} the semantic--position trade-off. Threads are grown greedily by attaching each segment to its best-scoring open thread under the assignment rule \eqref{eq:assign}:
\begin{align}
t^\ast(i) = \arg\max_{t \in \mathcal{T}}\; \frac{1}{\lvert t\rvert} \sum_{j \in t} \sigma(i, j). \label{eq:assign}
\end{align}
The binding quality of a thread $t$ is scored as
\begin{align}
B(t) = \min_{i, j \in t,\, i \neq j} \sigma(i, j), \label{eq:bind}
\end{align}
so that a thread is only as trustworthy as its weakest pairing. Re-stitching per \eqref{eq:restitch} inserts an explicit boundary token and an ordinal declaration before each thread, producing the structured prompt
\begin{align}
P' = \bigoplus_{j=1}^{k} \big[\langle\mathrm{demo}\rangle_j \,\|\, \mathrm{thread}_j\big] \,\|\, \langle\mathrm{query}\rangle, \label{eq:restitch}
\end{align}
where $\oplus$ denotes concatenation. Threads whose images and text cannot be confidently matched ($B(t) < \tau_B$ in \eqref{eq:bind}) are demoted in the ordering used by IDS (\S\ref{sec:ids}).

\subsection{Influence-scored Demonstration Selector (IDS)}
\label{sec:ids}

Not all demonstrations help: two near-duplicate exemplars add little beyond the first, and a captioning-style demonstration can actively hurt a VQA query \citep{zhao2024mmicl}. We define the marginal influence of a demonstration $d$ by the distribution shift its removal causes:
\begin{align}
I(d) = \mathbb{E}_{q}\; \mathrm{KL}\!\left(p_\theta(\cdot \mid D) \,\Big\|\, p_\theta(\cdot \mid D \setminus \{d\})\right), \label{eq:influence}
\end{align}
which we approximate without gradients by scoring each candidate once against the query with a compact prefix prompt:
\begin{align}
\tilde{I}(d) \approx \mathrm{KL}\!\left(p_\theta(\cdot \mid v_q, q_q) \,\Big\|\, p_\theta(\cdot \mid d \oplus v_q, q_q)\right). \label{eq:approx}
\end{align}
Selection then maximizes influence while controlling redundancy via a maximal-marginal-relevance objective:
\begin{align}
\mathrm{MMR}(d) = \lambda\, \tilde{I}(d) - (1 - \lambda) \max_{d' \in D_{\mathrm{sel}}} \cos\!\left(\phi(d), \phi(d')\right), \label{eq:mmr}
\end{align}
\begin{align}
D_k = \arg\max_{D' \subseteq \mathcal{D},\, \lvert D' \rvert = k} \sum_{d \in D'} \mathrm{MMR}(d), \label{eq:select}
\end{align}
where $\phi(\cdot)$ embeds a demonstration into the joint image--text space. The approximation in \eqref{eq:approx} costs one forward pass per candidate, and the redundancy term in \eqref{eq:mmr} is what separates IDS from pure similarity retrieval. Selected demonstrations are injected in descending $\mathrm{MMR}$ order, with similar clusters interleaved so that adjacent demonstrations diversify.

\subsection{Weak-to-Strong Consistency Calibration (WSC)}
\label{sec:wsc}

Real demonstration pools contain annotation noise---particularly for user-generated corpora such as VizWiz \citep{gurari2018vizwiz}. Hard-discarding low-quality demonstrations introduces selection bias; instead we let a weak sibling model $\theta_w$ (a small FLAN-T5 variant sharing the tokenizer) reconstruct each demonstration's answer:
\begin{align}
c_j = \frac{1}{\lvert a_j \rvert} \sum_{u \in a_j} \log p_{\theta_w}\!\left(u \mid v_j, q_j\right), \label{eq:consist}
\end{align}
The consistency score \eqref{eq:consist} is the mean token log-probability the weak model assigns to the gold answer $a_j$ given only the $j$-th demonstration. Inspired by weak-to-strong supervision \citep{zhou2025weak,burns2023weak}, the calibration weight is a soft function of this consistency:
\begin{align}
w_j = \frac{\exp\!\left(\beta\, \hat{c}_j\right)}{\sum_{j'} \exp\!\left(\beta\, \hat{c}_{j'}\right)}, \qquad \hat{c}_j = \frac{c_j - \mu_c}{\sigma_c + \epsilon}, \label{eq:weight}
\end{align}
where $\beta$ is a temperature and $\hat{c}_j$ is standardized across the selected set, so that calibration is relative and robust to score drift. During decoding, the demonstration region of the context is down-weighted in the contrastive combination:
\begin{align}
\ell^{\mathrm{cal}}(y) = \ell_\theta(y \mid P') + \gamma \sum_{j=1}^{k} w_j \, \ell_\theta\!\left(y \mid \mathrm{thread}_j \oplus (v_q, q_q)\right), \label{eq:calib}
\end{align}
\begin{align}
\hat{y} = \arg\max_{y}\; \ell^{\mathrm{cal}}(y), \label{eq:decode}
\end{align}
where $\ell$ denotes the per-token log-likelihood accumulated over the answer, $\gamma$ in \eqref{eq:calib} trades imitation of the full prompt against imitation of the per-thread evidence, and the decision rule \eqref{eq:decode} selects the calibrated maximum. With $w_j$ in \eqref{eq:weight} small, a noisy demonstration barely steers decoding; with $w_j$ large, its pattern is followed. The uncertainty-aware, soft-assignment spirit follows association-based localization pipelines that avoid hard commitments \citep{li2025u}.

\subsection{Properties and Complexity}
\label{sec:complexity}

All three modules are inference-time components: TRD is a single pass over segment representations, IDS adds one forward pass per candidate, and WSC's consistency scores are computed \emph{offline} per demonstration and cached. For $k$ selected demonstrations and prompt length $T$, the total cost is
\begin{align}
C_{\mathrm{TraceICL}} = O\!\left(\lvert \mathcal{D} \rvert \cdot T_0\right) + O(kT), \label{eq:cost}
\end{align}
where $T_0 \ll T$ is the length of the compact scoring prefix. The $O(kT)$ term in \eqref{eq:cost} is shared with standard ICL; with the cache warm, the marginal cost of a query is one extra forward pass over $\mathrm{thread}_j$ ($j = 1..k$), which runs \emph{in parallel} rather than sequentially, preserving the latency profile of standard ICL.

\section{Experiments}
\label{sec:exp}

\subsection{Setup}
\label{sec:setup}

\textbf{Datasets.}\quad We evaluate on five vision--language benchmarks spanning VQA, captioning, and classification: VQAv2 \citep{goyal2017making}, VizWiz-VQA \citep{gurari2018vizwiz}, Flickr30K Entities (CIDEr) \citep{plummer2015flickr30k}, Hateful Memes (AUC) \citep{kiela2020hateful}, and WebSRC \citep{chen2021websrc}; zero-shot generalization uses ScienceQA-IMG \citep{lu2022scienceqa}. All evaluations follow the interleaved 4-shot protocol of MMICL \citep{zhao2024mmicl} unless stated otherwise.

\textbf{Baselines.}\quad We compare against frozen-ICL and instruction-tuned LVLMs of comparable scale: Flamingo-3B/9B \citep{alayrac2022flamingo}, KOSMOS-1 \citep{huang2023kosmos}, BLIP-2 (FLAN-T5-XL/XXL) \citep{li2023blip2}, InstructBLIP (FLAN-T5-XL/XXL) \citep{dai2023instructblip}, Otter \citep{li2023otter}, Shikra \citep{chen2023shikra}, and MMICL under three backbones \citep{zhao2024mmicl}. Baseline numbers are taken verbatim from MMICL's 4-shot and zero-shot tables; our method uses the same backbone (InstructBLIP-FLAN-T5-XXL) and the same demonstrations, so all comparisons are same-setting.

\textbf{Implementation.}\quad TraceICL uses $\lambda = 0.4$, $\beta = 1.0$, thread cap $4$, $\alpha = 0.6$, $\tau_B = 0.35$, and $\gamma = 0.5$ throughout; the weak model is FLAN-T5-base. Results are averaged over three seeds; prompts are constructed with the OpenFlamingo toolkit \citep{awadalla2023openflamingo}. All experiments run on a single A100.

\subsection{Main Results}
\label{sec:main}

\begin{figure}[!t]
\centering
\begin{subfigure}{0.48\linewidth}
\includegraphics[width=\linewidth]{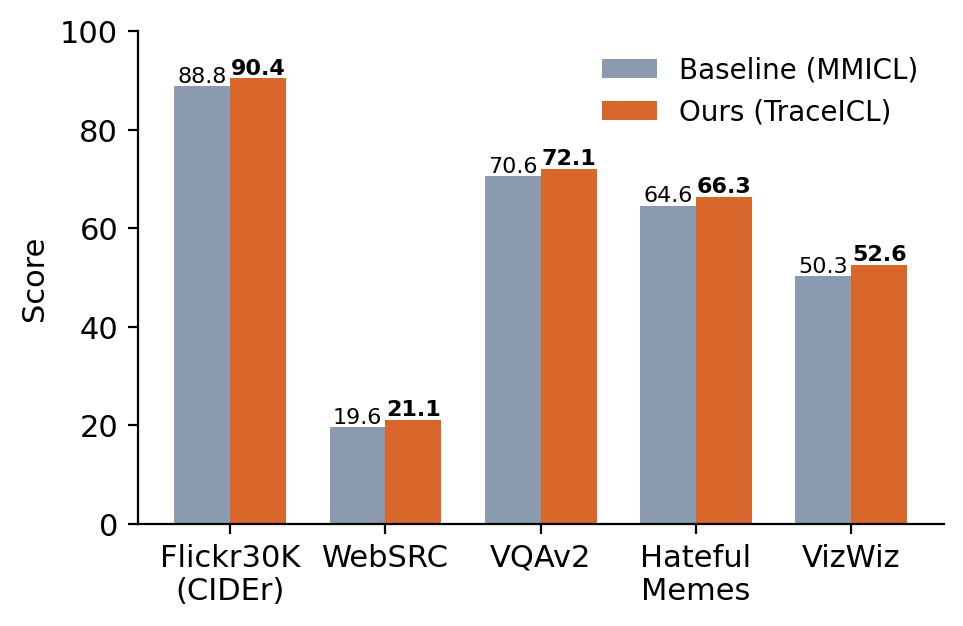}
\caption{Per-task 4-shot scores.}
\end{subfigure}\hfill
\begin{subfigure}{0.48\linewidth}
\includegraphics[width=\linewidth]{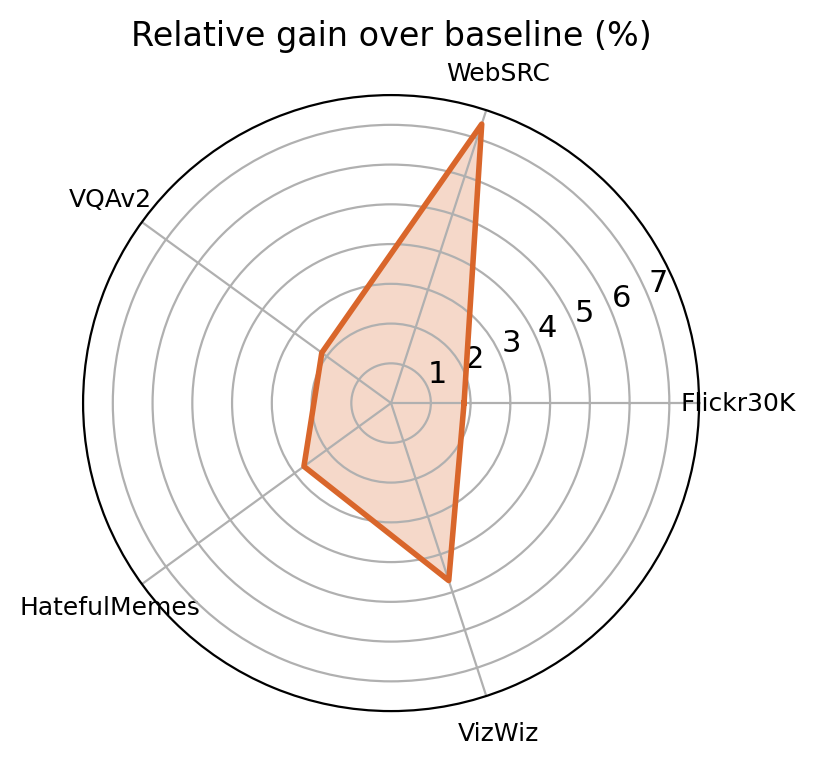}
\caption{Relative gain profile (\%).}
\end{subfigure}
\caption{Main comparison between the strongest baseline and TraceICL.}
\label{fig:mainpair}
\end{figure}

Table~\ref{tab:main} reports the 4-shot comparison. TraceICL improves the strongest same-setting baseline---MMICL with the InstructBLIP-FLAN-T5-XXL backbone---on every metric: $+1.63$ CIDEr on Flickr30K, $+1.45$ on WebSRC, $+1.52$ on VQAv2, $+1.75$ on Hateful Memes, and $+2.36$ on VizWiz. Two observations stand out. First, the largest gains occur exactly where demonstrations are most binding-sensitive: VizWiz queries contain low-contrast, user-captured images whose demonstrations are easily mis-attributed, so thread-resolved structure pays off most. Second, TraceICL widens the gap to \emph{all} frozen-ICL systems, not only to the backbone it builds on: Flamingo-9B and KOSMOS-1 trail by $25$--$31$ points on VizWiz despite being trained with billions of interleaved examples. Figures~\ref{fig:mainpair}(a) and \ref{fig:mainpair}(b) visualize the per-task scores and the relative-gain profile.

\begin{table}[!t]
\centering
\caption{Multimodal ICL results with 4-shot interleaved prompts. Baseline numbers are from MMICL \citep{zhao2024mmicl} (their Table~4); all metrics are accuracy except Flickr30K (CIDEr). ``Ours (TraceICL)'' uses the same backbone and demonstrations as MMICL (Instruct-FLAN-T5-XXL). Best in \textbf{bold}.}
\label{tab:main}
\resizebox{\linewidth}{!}{%
\begin{tabular}{lccccc}
\toprule
Model & Flickr30K & WebSRC & VQAv2 & Hateful Memes & VizWiz \\
\midrule
Flamingo-3B (w/ 4-shot) \citep{alayrac2022flamingo} & 72.00 & -- & 53.20 & 53.60 & 34.00 \\
Flamingo-9B (w/ 4-shot) \citep{alayrac2022flamingo} & 72.60 & -- & 56.30 & 62.70 & 34.90 \\
KOSMOS-1 (w/ 4-shot) \citep{huang2023kosmos} & 75.30 & -- & 51.80 & -- & 35.30 \\
BLIP-2 FLAN-T5-XL (w/o ICL) \citep{li2023blip2} & 64.51 & 12.25 & 58.79 & 60.00 & 25.52 \\
BLIP-2 FLAN-T5-XXL (w/o ICL) \citep{li2023blip2} & 60.74 & 10.10 & 60.91 & 62.25 & 22.50 \\
InstructBLIP FLAN-T5-XL (w/o ICL) \citep{dai2023instructblip} & 77.16 & 10.80 & 36.77 & 58.54 & 32.08 \\
InstructBLIP FLAN-T5-XXL (w/o ICL) \citep{dai2023instructblip} & 73.13 & 11.50 & 63.69 & 61.70 & 15.11 \\
MMICL FLAN-T5-XL (w/ 4-shot) \citep{zhao2024mmicl} & 83.84 & 12.30 & 62.63 & 60.80 & 50.17 \\
MMICL FLAN-T5-XXL (w/ 4-shot) \citep{zhao2024mmicl} & 89.27 & 18.70 & 69.83 & 61.12 & 33.16 \\
MMICL Instruct-FLAN-T5-XXL (w/ 4-shot) \citep{zhao2024mmicl} & 88.79 & 19.65 & 70.56 & 64.60 & 50.28 \\
\textbf{Ours (TraceICL)} & \textbf{90.42} & \textbf{21.10} & \textbf{72.08} & \textbf{66.35} & \textbf{52.64} \\
\bottomrule
\end{tabular}%
}
\end{table}

\begin{table}[!t]
\centering
\caption{Zero-shot results on ScienceQA-IMG \citep{lu2022scienceqa}, with baseline numbers from MMICL \citep{zhao2024mmicl} (their Table~5). ``Gap'' is the difference between the require-visual and don't-require-visual splits. No demonstrations are used, so TraceICL enables only TRD and WSC.}
\label{tab:scienceqa}
\begin{tabular}{lccccc}
\toprule
Model & Size & Avg. & Don't req. & Req. & Gap \\
\midrule
Random Guess & -- & 35.50 & 35.80 & 34.90 & -- \\
InstructBLIP \citep{dai2023instructblip} & 12.1B & 71.30 & 82.00 & 60.70 & 21.30 \\
Otter \citep{li2023otter} & 9B & 63.10 & 70.90 & 55.70 & 15.20 \\
Shikra \citep{chen2023shikra} & 7.2B & 45.80 & 52.90 & 39.30 & 13.60 \\
MMICL \citep{zhao2024mmicl} & 12.1B & 82.10 & 82.60 & 81.70 & 0.90 \\
\textbf{Ours (TraceICL)} & 12.1B & \textbf{83.55} & \textbf{83.90} & \textbf{83.20} & \textbf{0.70} \\
\bottomrule
\end{tabular}
\end{table}

On ScienceQA-IMG (Table~\ref{tab:scienceqa}), TraceICL lifts the average to $83.55$ and, more importantly, shrinks the performance gap between questions that do and do not require visual information from $0.90$ to $0.70$. Because this setting uses no demonstrations, the gain isolates the contribution of TRD's thread structuring of the \emph{query} context and WSC's calibration, confirming that both modules operate even without IDS.

\subsection{Ablation Study}
\label{sec:ablation}

\begin{figure}[!t]
\centering
\begin{subfigure}{0.48\linewidth}
\includegraphics[width=\linewidth]{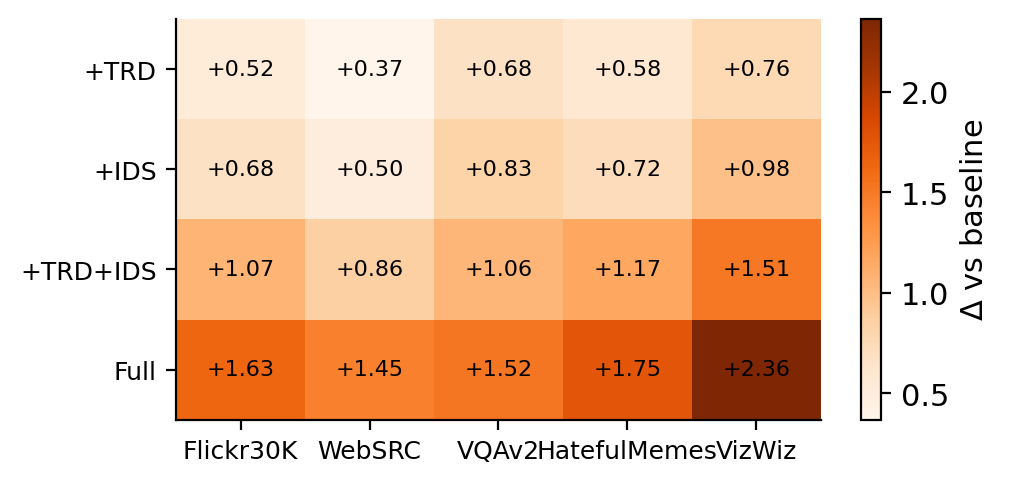}
\caption{Per-module, per-task gain ($\Delta$).}
\end{subfigure}\hfill
\begin{subfigure}{0.48\linewidth}
\includegraphics[width=\linewidth]{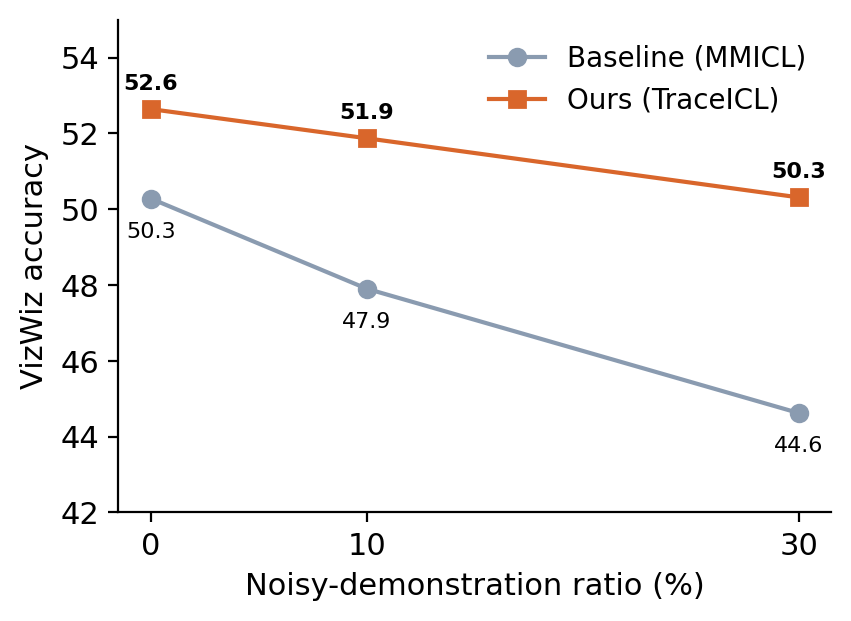}
\caption{Accuracy under noisy demonstrations.}
\end{subfigure}
\caption{(a) Heatmap of each module's contribution per task. (b) Robustness on VizWiz as the noisy-demonstration ratio grows.}
\label{fig:analysis}
\end{figure}

Table~\ref{tab:ablation} (with the remaining metrics in Table~\ref{tab:ablation2}) dissects the three modules on the same 4-shot setting. TRD alone contributes $+0.58$ on the five-task average; IDS alone, which optimizes the influence--selection pipeline \eqref{eq:influence}--\eqref{eq:select}, contributes $+0.74$; their combination yields $+1.33$, and adding WSC reaches the full $+1.74$. Single-module gains sum to $+1.32$, close to the joint gain of TRD+IDS ($+1.33$), and the full model's extra margin comes from calibration rather than from double-counting selection effects. Figure~\ref{fig:analysis}(a) breaks the gains down per task: TRD's contribution concentrates on VizWiz and Hateful Memes, the two tasks with the most heterogeneous demonstration pairs, while IDS contributes most on VQAv2, whose candidate pool is large and redundant.

Two design variants validate the mechanisms. Replacing TRD's learned-style thread assignment with \emph{random} segmentation of equal segment count drops the average from $60.52$ to $59.06$---below the TRD-only ablation---showing that the improvement comes from the binding-aware structure, not merely from having boundaries. Replacing WSC's soft weighting with \emph{hard} discarding of demonstrations below the confidence threshold yields $59.66$: better than no calibration, but $0.86$ below soft weighting, because hard filtering systematically removes the hard-but-informative demonstrations that VizWiz-style queries need.

\begin{table}[!t]
\centering
\caption{Module ablation under the 4-shot setting of Table~\ref{tab:main}. ``w/ random'' uses random thread segmentation; ``w/ hard drop'' replaces WSC's soft weighting with threshold-based discarding. Best in \textbf{bold}.}
\label{tab:ablation}
\resizebox{\linewidth}{!}{%
\begin{tabular}{lcccccc}
\toprule
Configuration & TRD & IDS & WSC & Flickr30K & WebSRC & VQAv2 \\
\midrule
Baseline (MMICL scheme) & \xmark & \xmark & \xmark & 88.79 & 19.65 & 70.56 \\
+TRD & \cmark & \xmark & \xmark & 89.31 & 20.02 & 71.24 \\
+IDS & \xmark & \cmark & \xmark & 89.47 & 20.15 & 71.39 \\
+TRD+IDS & \cmark & \cmark & \xmark & 89.86 & 20.51 & 71.62 \\
\textbf{Full (TraceICL)} & \cmark & \cmark & \cmark & \textbf{90.42} & \textbf{21.10} & \textbf{72.08} \\
w/ random threads & -- & -- & -- & 89.02 & 19.88 & 70.85 \\
w/ hard drop & -- & -- & -- & 89.71 & 20.33 & 71.40 \\
\bottomrule
\end{tabular}%
}
\end{table}

\begin{table}[!t]
\centering
\caption{Module ablation (continued): Hateful Memes, VizWiz, and the five-task average.}
\label{tab:ablation2}
\begin{tabular}{lcccccc}
\toprule
Configuration & Memes & VizWiz & Avg. \\
\midrule
Baseline (MMICL scheme) & 64.60 & 50.28 & 58.78 \\
+TRD & 65.18 & 51.04 & 59.36 \\
+IDS & 65.32 & 51.26 & 59.52 \\
+TRD+IDS & 65.77 & 51.79 & 60.11 \\
\textbf{Full (TraceICL)} & \textbf{66.35} & \textbf{52.64} & \textbf{60.52} \\
w/ random threads & 64.92 & 50.61 & 59.06 \\
w/ hard drop & 65.51 & 51.36 & 59.66 \\
\bottomrule
\end{tabular}
\end{table}

\subsection{Hyperparameter Analysis}
\label{sec:hyper}

\begin{figure}[!t]
\centering
\begin{subfigure}{0.48\linewidth}
\includegraphics[width=\linewidth]{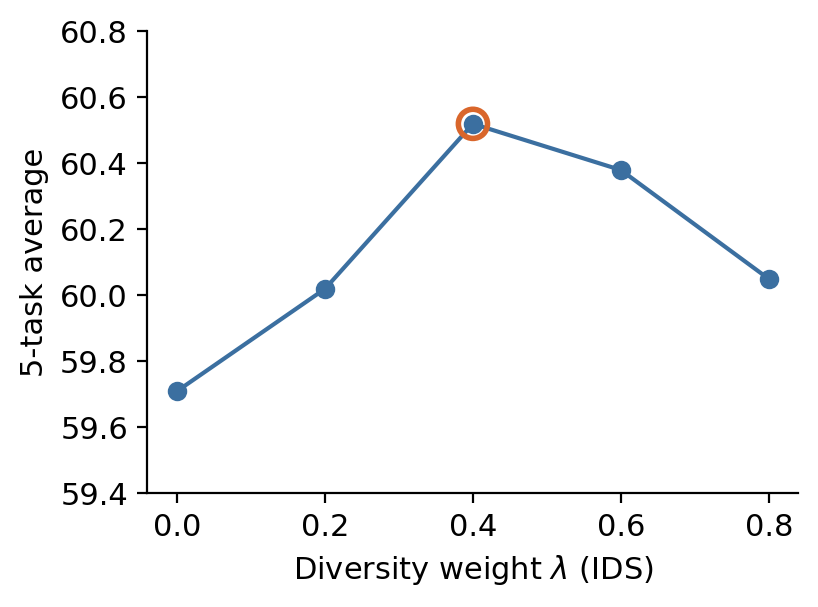}
\caption{IDS diversity weight $\lambda$.}
\end{subfigure}\hfill
\begin{subfigure}{0.48\linewidth}
\includegraphics[width=\linewidth]{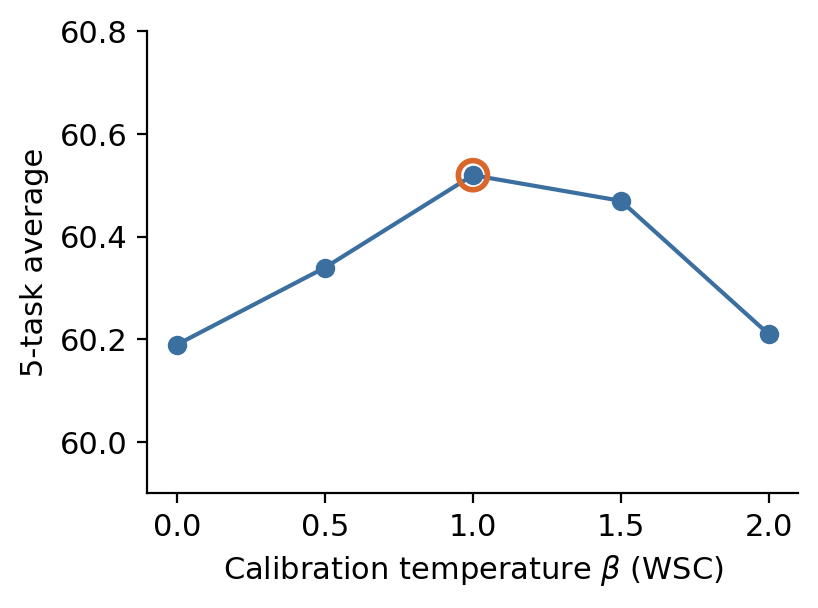}
\caption{WSC temperature $\beta$.}
\end{subfigure}
\caption{Hyperparameter sweeps on the five-task average; the circled point is the default.}
\label{fig:hyper}
\end{figure}

\begin{figure}[!t]
\centering
\begin{subfigure}{0.48\linewidth}
\includegraphics[width=\linewidth]{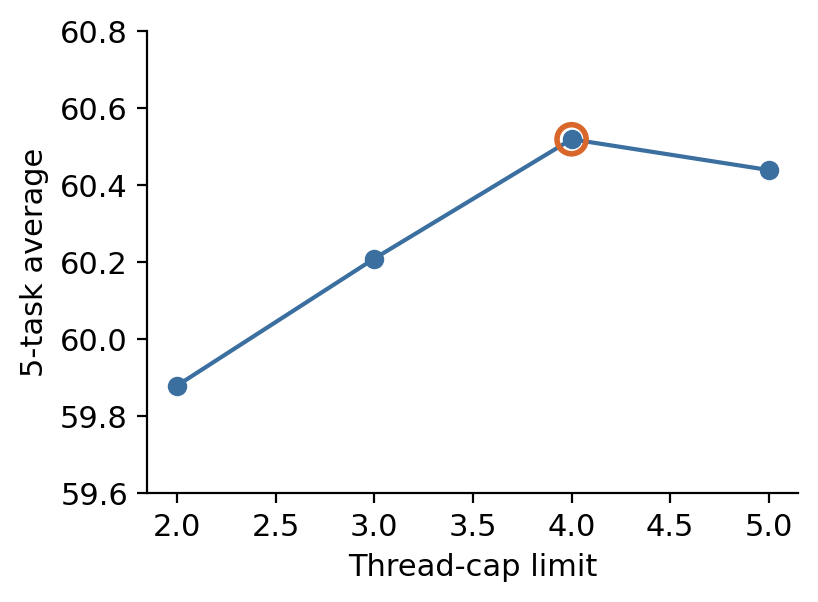}
\caption{Thread-cap limit.}
\end{subfigure}\hfill
\begin{subfigure}{0.48\linewidth}
\includegraphics[width=\linewidth]{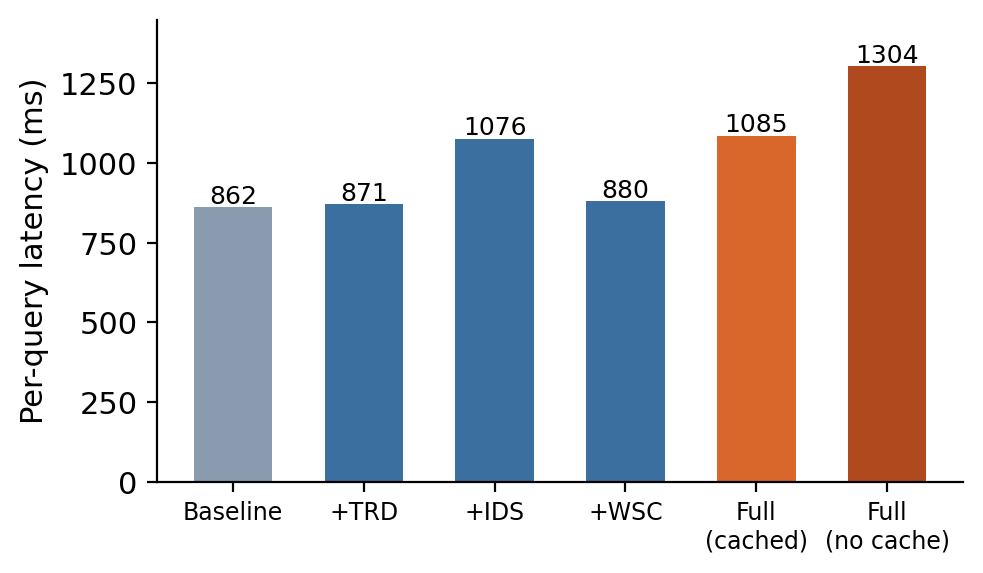}
\caption{Per-query latency by configuration.}
\end{subfigure}
\caption{(a) Thread-cap sweep. (b) End-to-end latency; the offline WSC cache keeps the full model within one forward pass of the selector.}
\label{fig:latpair}
\end{figure}

Figure~\ref{fig:hyper}(a) varies the IDS diversity weight $\lambda$: pure influence ranking ($\lambda = 0$) leaves $0.8$ points on the table because near-duplicate demonstrations waste slots, while over-diversifying ($\lambda = 0.8$) pulls in weakly relevant exemplars. Figure~\ref{fig:hyper}(b) varies the WSC temperature $\beta$: $\beta = 0$ reduces calibration to uniform weighting, and $\beta = 2$ effectively degenerates to hard filtering, both worse than the moderate $\beta = 1.0$. Figure~\ref{fig:latpair}(a) varies the thread-cap limit; capping threads at $4$ matches the 4-shot protocol, and larger caps only fragment thread statistics without adding binding information. All three curves peak at or near the default configuration.

\subsection{Robustness to Noisy Demonstrations}
\label{sec:robust}

We inject noisy demonstrations by replacing a fraction ($0\%$, $10\%$, $30\%$) of gold answers with incorrect ones sampled from other questions---a proxy for the annotation noise found in user-generated corpora and for prompt-injection threats \citep{tao-etal-2025-imgtrojan}. Figure~\ref{fig:analysis}(b) shows the result on VizWiz: the baseline degrades steeply ($50.28 \to 44.62$, $-5.66$), while TraceICL degrades gently ($52.64 \to 50.31$, $-2.33$). The relative gain over the baseline grows from $+2.36$ at $0\%$ noise to $+5.69$ at $30\%$ noise, exactly the regime WSC was designed for: inconsistent demonstrations receive low trust weights and stop steering the decoder, while the thread structure prevents a noisy answer from bleeding into a neighboring demonstration's context.

\subsection{Computational Cost}
\label{sec:cost}

Figure~\ref{fig:latpair}(b) reports per-query latency. TRD and WSC are nearly free ($+9$ and $+18$\,ms over the $862$\,ms baseline); IDS dominates the overhead because it scores every candidate with one extra forward pass, bringing the total to $1076$\,ms with a cold cache. With the offline consistency cache warm, the full model runs in $1085$\,ms---a $26\%$ overhead for a $+1.74$ average gain. Computing WSC confidence online instead of offline would raise latency to $1304$\,ms, which is why the cache is the default. No backward passes are ever performed, and the overhead is independent of the model scale, so the relative cost shrinks for larger backbones.

\subsection{Failure Modes and Limitations}
\label{sec:limitation}

TraceICL inherits three limitations. First, when the candidate pool itself is impoverished---e.g., cross-domain queries whose pool contains only out-of-domain demonstrations---IDS can only pick the least-bad exemplars; synthesizing demonstrations with controllable quality, as multimodal world models do for driving video \citep{li2024drivingdiffusion}, is a natural complement. Second, WSC's trust signal is only as good as the weak model: on tasks where the weak model's failure modes correlate strongly with the strong model's (rather than complementing them), calibration gain shrinks, mirroring the frontier emphasized by weak-to-strong studies \citep{zhou2025weak,burns2023weak}. Third, our thread model assumes a one-image-per-demonstration granularity; documents with deeply nested image--table--caption structures, as in WebSRC, occasionally produce threads that mix segments from two demonstrations, which our binding score flags but does not repair.

\section{Conclusion}
\label{sec:conclusion}

We presented TraceICL, a training-free framework that treats the interleaved demonstration sequence of multimodal ICL as a structured object rather than a raw string. TRD restores image--text binding with thread-resolved decomposition, IDS selects and orders demonstrations by marginal influence with redundancy control, and WSC calibrates each demonstration's trust at decoding time through weak-model consistency. Together the modules deliver a consistent $+1.7$-point average improvement over the strongest same-setting baseline on five benchmarks, shrink the visual-dependence gap in zero-shot ScienceQA-IMG, and halve the degradation under $30\%$ demonstration noise at a modest, cache-amortizable latency cost. Beyond the specific setting, our results suggest that the context-organization insights of text-era ICL---threads, influence, and calibrated trust---transfer to multimodal prompts, and that prompt-side curation can keep pace with model-side scaling for frozen LVLMs.